\documentclass[lettersize,journal]{IEEEtran}
\usepackage{amsmath,amsfonts}
\usepackage{algorithmic}
\usepackage{algorithm}
\usepackage{array}
\usepackage{arydshln}
\usepackage[caption=false,font=normalsize,labelfont=sf,textfont=sf]{subfig}
\usepackage{textcomp}
\usepackage{stfloats}
\usepackage{url}
\usepackage{verbatim}
\usepackage{graphicx}
\usepackage{cite}
\usepackage{bm}
\begin{document}

\title{MSCT: Addressing Time-Varying Confounding with Marginal Structural Causal Transformer for Counterfactual Post-Crash Traffic Prediction}

\author{Shuang~Li,
        Ziyuan~Pu,~\IEEEmembership{Member,~IEEE,}
        Nan~Zhang,
        Duxin~Chen,~\IEEEmembership{Member,~IEEE,}
        Lu~Dong,~\IEEEmembership{Member,~IEEE,}
        Daniel~J.~Graham,
        Yinhai~Wang,~\IEEEmembership{Fellow,~IEEE,}
        
\thanks{This work has been supported by the Postgraduate Research\&Practice Innovation Program of Jiangsu Province (project number: KYCX22\_0271) and the Ministry of Transport of PRC Key Laboratory of Transport Industry of Comprehensive Transportation Theory (Nanjing Modern Multimodal Transportation Laboratory) (project number: MTF2023002)(Crossponding author: Ziyuan Pu)}
\thanks{Shuang Li are with School of Transportation, Southeast University, Nanjing, 211189, China, and Department of Civil and Environmental Engineering, Imperial College London, London, SW7 2AZ, UK (shuangli\_seu@seu.edu.cn)}
\thanks{Ziyuan Pu is with School of Transportation, Southeast University, Nanjing, 211189, China (ziyuanpu@seu.edu.cn)}
\thanks{Daniel J. Graham, and Nan Zhang are with Department of Civil and Environmental Engineering, Imperial College London, London, SW7 2AZ, UK (d.j.graham@imperial.ac.uk; nan.zhang16@imperial.ac.uk)}
\thanks{Duxin Chen is with the School of Mathematics, Southeast University, Nanjing 211189, China(chendx@seu.edu.cn)}
\thanks{Lu Dong is with the School of Cyber Science and Engineering, Southeast University, Nanjing 211189, China(ldong90@seu.edu.cn)}
\thanks{Yinhai Wang is with the Department of Civil and Environmental Engineering, University of Washington, Seattle, WA 98195 USA (yinhai@uw.edu)}}

\markboth{Journal of \LaTeX\ Class Files}%
{Shell \MakeLowercase{\textit{et al.}}: A Sample Article Using IEEEtran.cls for IEEE Journals}

\maketitle

\vspace{-1cm}

\begin{abstract}
Traffic crashes profoundly impede traffic efficiency and pose economic challenges. Accurate prediction of post-crash traffic status provides essential information for evaluating traffic perturbations and developing effective solutions. Previous studies have established a series of deep learning models to predict post-crash traffic conditions, however, these correlation-based methods cannot accommodate the biases caused by time-varying confounders and the heterogeneous effects of crashes. The post-crash traffic prediction model needs to estimate the counterfactual traffic speed response to hypothetical crashes under various conditions, which demonstrates the necessity of understanding the causal relationship between traffic factors. Therefore, this paper presents the Marginal Structural Causal Transformer (MSCT), a novel deep learning model designed for counterfactual post-crash traffic prediction. To address the issue of time-varying confounding bias, MSCT incorporates a structure inspired by Marginal Structural Models and introduces a balanced loss function to facilitate learning of invariant causal features. The proposed model is treatment-aware, with a specific focus on comprehending and predicting traffic speed under hypothetical crash intervention strategies. In the absence of ground-truth data, a synthetic data generation procedure is proposed to emulate the causal mechanism between traffic speed, crashes, and covariates. The model is validated using both synthetic and real-world data, demonstrating that MSCT outperforms state-of-the-art models in multi-step-ahead prediction performance. This study also systematically analyzes the impact of time-varying confounding bias and dataset distribution on model performance, contributing valuable insights into counterfactual prediction for intelligent transportation systems.
\end{abstract}

\vspace{-0.2cm}
\section{Introduction}
\IEEEPARstart{T}{raffic} crashes always cause severe travel delays and substantial economic losses. According to the World Health Organisation (WHO), the cost of road traffic injuries in the global macroeconomy is estimated to be 1.8 trillion dollars \cite{WHO2023}. In addition,  the risk of crashes during post-crash duration is six times greater than that of primary crashes \cite{tedesco1994development}. The ability to accurately predict post-crash traffic conditions is essential for Intelligent Transportation Systems (ITS), not only to help traffic managers improve traffic management and control strategies but also to ensure the safety of road users.

\begin{figure}[t]
\centering
\setlength{\abovecaptionskip}{0.cm}
\hspace{-0.3in}
\includegraphics[width=3.5 in]{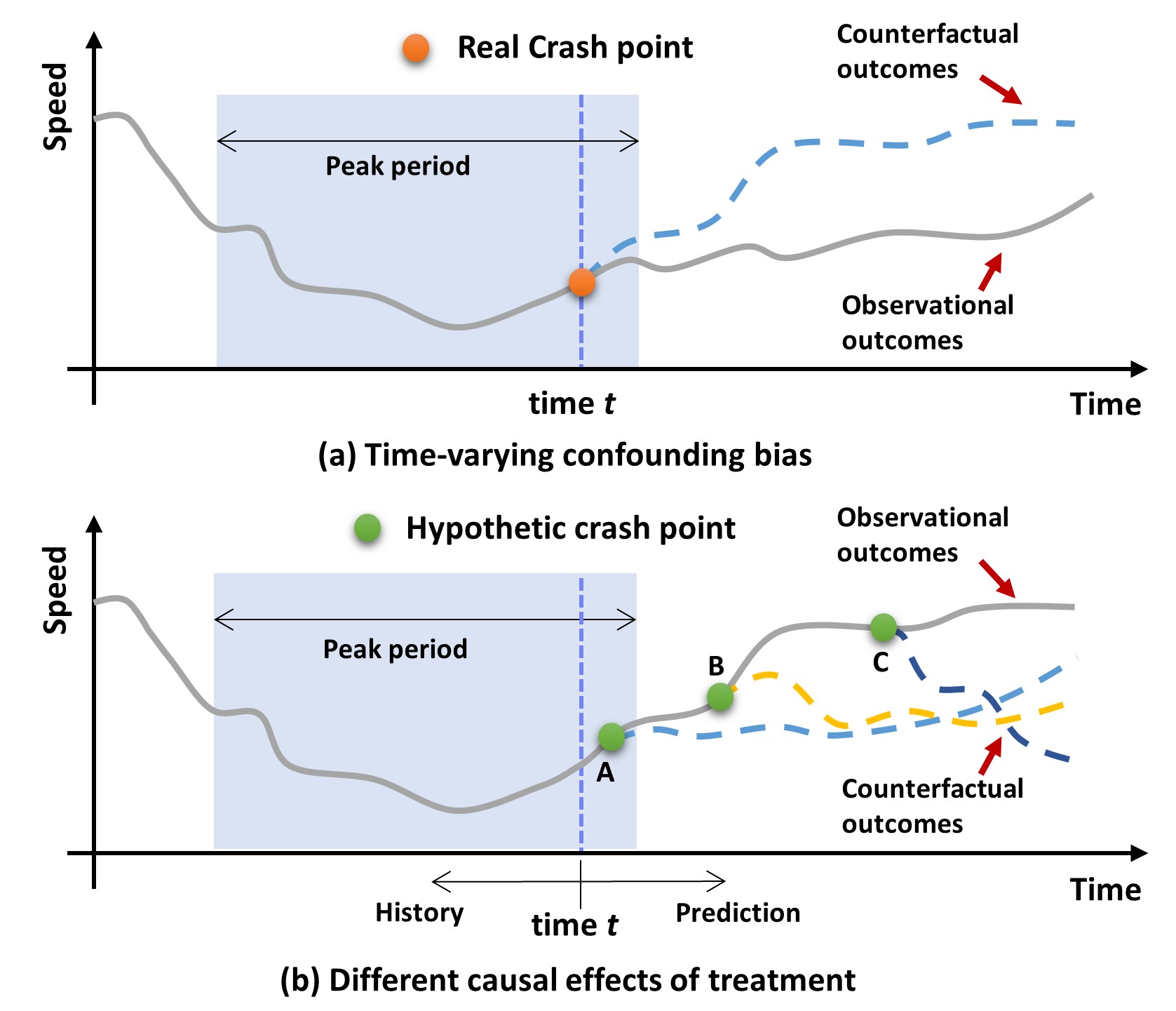}
\caption{Illustration of counterfactual traffic speed under the impact of crashes.}
\label{fig_1}
\end{figure}

Post-crash traffic prediction is a counterfactual prediction task that requires the development of a treatment-aware model, enabling it to predict the traffic conditions under hypothetical intervention (crash) based on observational data \cite{pearl_causal_2009}. However, many existing machine-learning models are data-driven, fundamentally focusing on learning correlations within the data rather than establishing causal relationships. Since "correlation does not imply causation," data-driven methods often overlook selection biases introduced by confounders—variables that causally influence both the treatment and the outcomes. This oversight can lead to unforeseen biases and increased generalization errors when predicting counterfactual traffic conditions \cite{alaa_limits_2018}.

Fig.\ref{fig_1} illustrates the problem of post-crash traffic prediction. As shown in Fig.\ref{fig_1}(a), at the end of the peak period, a real crash occurred at time \textit{t}. Following the crash, there was a slight increase in speed (solid line). Data-driven models trained on this observed data might mistakenly learn that crashes contribute to a slight speed increase. However, the counterfactual speed (dotted line) would have been higher if no crash had occurred. Here, the \textit{time} acts as a time-varying confounder, potentially preventing models from accurately capturing the true temporal causal mechanism. Fig. \ref{fig_1}(b) shows the heterogeneous effects of hypothetical crashes at different time points. The prediction model needs to estimate the speed after a crash that did not actually occur, thereby highlighting the imperative to comprehend the causal effects of crashes across diverse circumstances. Therefore, post-crash traffic prediction should be considered as a counterfactual prediction task, which means estimating the outcomes response to hypothetical intervention or treatment in various conditions. 

Based on causal theory, the crash indicator and the post-crash traffic speed are usually defined as \textit{Treatment} and \textit{Outcome}. To address post-crash traffic prediction, three major challenges remain to be solved: 
 \textit{1) The issue of selection biases in observational data, particularly the time-varying confounding biases resulting from time-varying confounders, needs to be addressed. \cite{mannering_selectivity_1986,karwa_causal_2011}; 2) Counterfactual prediction should be treatment-aware, namely make a causal estimation given the hypothetical intervention under different circumstances; and 3) Due to the unobservable nature of counterfactual outcomes, there is no ground truth to validate the prediction performance}.
 
In previous research, a multitude of factors have been demonstrated to impact traffic crashes and post-crash conditions significantly, and numerous Deep Learning (DL) models have been proposed for predicting post-crash traffic conditions \cite{miller_mining_2012, pan_forecasting_2015,wang_traffic_2016,xie_how_2019}. These studies adopt data-driven approaches to incorporate a comprehensive set of relevant variables into the model, leveraging the learning capabilities of DL for accurate predictions. However, these data-driven models rely on correlations rather than causation \cite{mannering_big_2020}, which may introduce unexpected errors when attempting to make causal estimations \cite{mannering_temporal_2018}.

Since randomized controlled trials (the best method for causal estimations) for traffic crashes are unfeasible and ethically untenable, the estimation from observational data has garnered considerable attention among scholars \cite{li_impacts_2013,li_quantifying_2016,mannering_unobserved_2016}. Traditional methods, such as propensity score matching and doubly robust, are mainly designed for single treatment rule and cross-section data. Considering traffic conditions are significantly affected by time-varying covariates, the methods designed for time-varying treatments, like Marginal Structural Models (MSMs) \cite{robins_marginal_2000} and G-estimation \cite{hernan_causal_2020}, are thus more suitable post-crash traffic speed prediction. However, these linear regression-based models are limited in fitting extensive traffic data and capturing long-term temporal dependencies \cite{mannering_big_2020}. 

Given the strong capabilities of DL technologies in handling large and complex data, the integration of DL networks with causal theory has emerged as a prevailing trend \cite{berrevoets_causal_2023}. Johansson et al.\cite{johansson_learning_2018} initially framed counterfactual inference as a domain adaptation problem and proposed a Balancing Neural Network (BNN) for estimating Individual Treatment Effects (ITE). Subsequently, various types of causal DL models have been established based on balancing rules and domain-invariant learning \cite{alaa_deep_2017, louizos_causal_2017,yoon_ganite_2018, shi_adapting_2019}. To predict the treatment response under time-varying confounding situations, several causal DL models have been proposed in the medical field, including recurrent marginal structural networks (RMSNs) \cite{lim_forecasting_2018}, counterfactual recurrent network (CRN) \cite{bica_estimating_2020}, G-Net \cite{li_g-net_2021}, and causal transformer (CT) \cite{melnychuk_causal_2022}. These models utilize the DL network to capture long-term temporal dependencies of patients' states. However, traffic conditions tend to exhibit more complex and dynamically fluctuating characteristics, and the occurrence of crashes is more difficult to predict than therapeutic strategies, making it more difficult to infer the evolution of post-crash conditions.

As it is impossible to simultaneously observe traffic states under both crash and non-crash conditions, a dataset with a known internal causal mechanism is necessary to evaluate the performance of counterfactual prediction. Despite the capability of traffic simulation software to produce data for diverse traffic scenarios, the intricate nature of scene modeling and parameter calibration poses challenges in comprehensively understanding causal mechanisms. Therefore, a Data Generation Process (DGP)
that directly reveal the causal relationship is necessary. However, the challenge of simulating traffic trends, such as the temporal dependency, peak and off-peak patterns, and the dissipation of congestion, remains an unresolved issue.

This study introduces a novel deep learning model, Marginal Structural Causal Transformer (MSCT), that aims to address the problem of counterfactual post-crash traffic prediction. Different from traditional DL prediction models that are mainly dedicated to fusing multi-source data and capturing spatial-temporal dependencies, our model additionally pays more focus on the causal relationship and is specifically tailored to address interventions related to traffic crashes. Our contributions are summarized as follows:

\begin{itemize}
    \item Inspired by causal theory, we develop a sequence-to-sequence causal deep learning model that applies the Transformer for counterfactual prediction. To deal with the time-varying confounding bias, an MSMs-inspired model structure is built, and the balanced loss is proposed to learn invariant features by domain generalized training procedure.
    \item The proposed DL model is treatment-aware, meaning it can predict heterogeneous traffic speed response to hypothetical crash intervention strategies. To the best of our knowledge, this is the first attempt to introduce the concept of treatment response prediction into traffic crash scenarios by deep learning techniques. 
    \item Due to the absence of ground-truth data, we propose a synthetic data generation procedure that can emulate traffic speed, considering the occurrence of traffic crashes. Our model is compared with state-of-the-art models using synthetic and real-world data, and it achieves superior performance. Systematic and empirical analyses have been conducted to investigate the impact of the degree of time-varying confounding bias and the imbalanced distribution of the dataset on model performance.
\end{itemize}

\section{Literature review}
\subsection{Traffic Speed Prediction under Incident Conditions}
Extensive methods have been proposed for estimating traffic speed. Early prediction models, such as ARIMA and VAR, are mainly parametric models. With the development of ML technologies, various models have been applied for speed prediction \cite{zhou_comprehensive_2022}. Recently, deep learning models have performed superiorly in capturing complicated and dynamic features of traffic conditions. A large amount of deep learning models for traffic speed prediction have emerged. These new models are mainly based on one or several combinations of state-of-the-art neural networks, such as RNN, LSTM, CNN, GCN, and Attention.

There are also several studies exploring the impact of incidents on traffic conditions. Miller et al. suggested a system to forecast the cost and effect of highway incidents using machine models \cite{miller_mining_2012}. Their system allows for predicting the duration of incidents and the cost of delay. Li and Chen collected the features of highway traffic conditions with non-recurrent congestion and developed a travel time prediction model by using MLP \cite{li_identifying_2013}. Pan et al. proposed a model to predict when and how travel time delays will occur in road networks due to traffic incidents \cite{pan_forecasting_2015}. Their model is built based on the attributes of historical cases, and a clustering algorithm is applied for new scenarios to determine initial propagation behavior. Javid et al. developed a framework to estimate travel time variability caused by traffic incidents by using a series of robust regression methods \cite{j_javid_framework_2018}. Xie et al. propose a deep learning model to quantify the impact of urban traffic incidents on traffic flows \cite{xie_how_2019}. The traffic incident data are processed by a classifier to extract latent features which then be introduced into the prediction model. Shafiei et al. built a simulation-based framework for incident impact analysis \cite{shafiei_short-term_2020}. ML models are applied in their framework for incident classification and demand prediction, which enhance traffic micro-simulation.

According to the existing literature, clustering or classifying the traffic incidents first is a common procedure for prediction, which helps recognize the impact level of a new incident. However, these methods are based on correlation of variables. When it comes to understanding the impact mechanism of traffic crashes, a causal relationship is necessary to avoid biased estimation. Therefore, different from the previous studies, this study aims to address the causal problem for post-crash traffic speed prediction and provide a more effective solution for multi-step-ahead prediction given the intervention strategies.

\subsection{Counterfactual Outcomes Prediction with Deep Learning}
In recent years, the DL community's interest in causality has greatly increased.  Traditionally, causal models have been primarily used for estimating average treatment effects in the policy research field, with the potential outcome framework (also known as Neyman-Rubin Causal Model) being a widely employed approach \cite{rubin_bayesian_1978,splawa-neyman_application_1990}. However, there is a growing interest in individual treatment effects (i.e., counterfactual outcomes), for which traditional linear regression-based models lack the capability to predict accurately. On the other hand, DL has contributed largely to prediction when applied to big data, but these models are black boxes and are more interested in correlations than causality between inputs and outputs, resulting in pool reliability for decision-making \cite{chakraborty_interpretability_2017}. To improve the ability of DL models to answer counterfactual questions, causal models have been widely combined for counterfactual outcome prediction.

The methods introducing causal models into DL are diverse. Johansson et al. combined domain adaptation and representation learning for causal effects inference \cite{johansson_learning_2018}. Alaa et al. built a DCN-PD model that utilizes a propensity-dropout regularization scheme to alleviate selection bias in the observational data \cite{alaa_deep_2017}. Louizos et al proposed CEVAE based on Variational Autoencoders which follows the causal structure of inference \cite{louizos_causal_2017}. Yoon et al. incorporated generative adversarial model architecture into individual treatment effect inference, which makes GAN popular in the field of causal estimation \cite{yoon_ganite_2018}. Li et al. proposed a conditional invariant deep domain generalization approach for domain-invariant representation learning. The gradient reversal is adopted to eliminate spurious correlations \cite{li_deep_2018}. Shi et al. developed a Dragonnet that applies propensity score to adjust the confounding bias \cite{shi_adapting_2019}. However, these models are designed for cross-sectional data, which cannot be applied for time-varying treatment response prediction.

Due to the greater importance of longitudinal data in the medical field, several scholars explored the adjustment of time-varying confounding bias by DL. Lim et al. first applied a recurrent neural network to predict the inverse probability of treatment weights (IPTWs), and built recurrent marginal structural networks (RMSN) \cite{lim_forecasting_2018}. Subsequently, Bica et al. proposed counterfactual recurrent networks (CRN) which apply domain adversarial training to balancing representations \cite{bica_estimating_2020}.  Li et al. designed a sequential DL framework for G-computation (G-Net) that enables counterfactual predictions under dynamic treatment strategies \cite{li_g-net_2021}. Melnychuk et al. also developed a domain adversarial representations-based model with multi-input transformer architecture (CT) \cite{melnychuk_causal_2022}. According to these state-of-art examples, two main ideas are applied for counterfactual outcomes prediction over time: one is designing a network to implement causal models, such as MSMs, and G-computation; the other is applying adversarial training to generate representations. To capture the temporal dependency, RNN, transformer, and sequence-to-sequence structure are the main methods. However, these models are designed for medical data, which possess distinct characteristics from transportation data. Therefore, to effectively adapt these models for transportation-related applications, we need to conduct more specific and tailored research to account for the time-varying treatment effects and confounding bias inherent in traffic data.

\subsection{Causal Traffic Prediction}
Causal effects estimating models have been widely applied in the traffic safety field to explore the causal effects of behavioral elements \cite{afghari_investigating_2022}, traffic policy \cite{li_impacts_2013}, road construction \cite{zhang_inferring_2022}, etc. These studies mainly focus on estimating average treatment effects which are useful to support traffic decision-making. Recently, causal machine learning methods have also applied to traffic fields, such as generalized random forest \cite{zhang_estimating_2021}, double machine learning \cite{liu_estimating_2022}, and doubly robust learning \cite{li2024inferring}, because of their outstanding performance in estimating heterogeneous treatment effects. These studies contributed greatly to solving the confounding bias of traffic data by causal models; however, they are mainly based on cross-sectional data.

Additionally, some researchers have explored the integration of causal discovery models into prediction models for traffic time-series data. For instance, Li et al. introduced a Lasso method that applies Granger causality to uncover potential dependencies among big data, leading to more robust traffic flow predictions \cite{li_robust_2015}. Similarly, Gao et al. integrated an iterative causal discovery algorithm into the temporal convolutional network, enhancing predictive model performance through the graph adjacency matrix of variables\cite{gao_gt-causin_2022}. Meanwhile, He et al. proposed a GNN-based framework, using a spatial-temporal Granger causality graph (STGC-GNNs) for traffic prediction \cite{he_stgc-gnns_2023}. While these studies apply causal models to extract causal graphs between variables, their primary focus remains on traditional prediction tasks rather than counterfactual outcomes prediction. As a result, these models may not be suitable for estimating traffic conditions under specific interventions, such as crashes, due to the potential selection bias present in observational data, which can lead to errors. Therefore, embedding confounding bias adjustment into deep learning models for time-series traffic data prediction becomes crucial for more accurate and reliable results.

\section{problem statement}
To estimate the causal effect of traffic crashes on speed over time, we describe the problem built upon the time-varying extension of the potential outcome framework[26]. Given that traffic data consists of long-term continuous time series, considering each road segment as an individual would result in excessively long observation periods. Therefore, to address this issue, we employ a specific length of time sequence to extract subsets of time series data based on traffic location information, creating multiple time blocks that serve as the unit. For each unit \textit{i} at time step \textit{t}, we have the following:  $d_x$-dimension time-varying covariates $X_{i,t} \in {\mathbb{R}^{{d_x}}}$ potential outcome (traffic speed) $Y_{i,t}$, and treatment (crash) ${T_{i,t}} \in \left[ {0,1} \right]$. If a crash occurs at time \textit{t} for unit \textit{i}, then $T_{i,t}=1$, otherwise, $T_{i,t}=0$. Further, each unit \textit{i} has $d_s$-dimension static features $S_i \in {\mathbb{R}^{d_s}}$, such as milepost, weather, and day of the week. Because we study the traffic states in 5-min intervals, the weather and day of the week are relatively static. For notational simplicity, we omit the subscript \textit{i} unless needed.

\begin{figure}[t]
    \centering
    \setlength{\abovecaptionskip}{0.cm}
    \includegraphics[width=3.5 in]{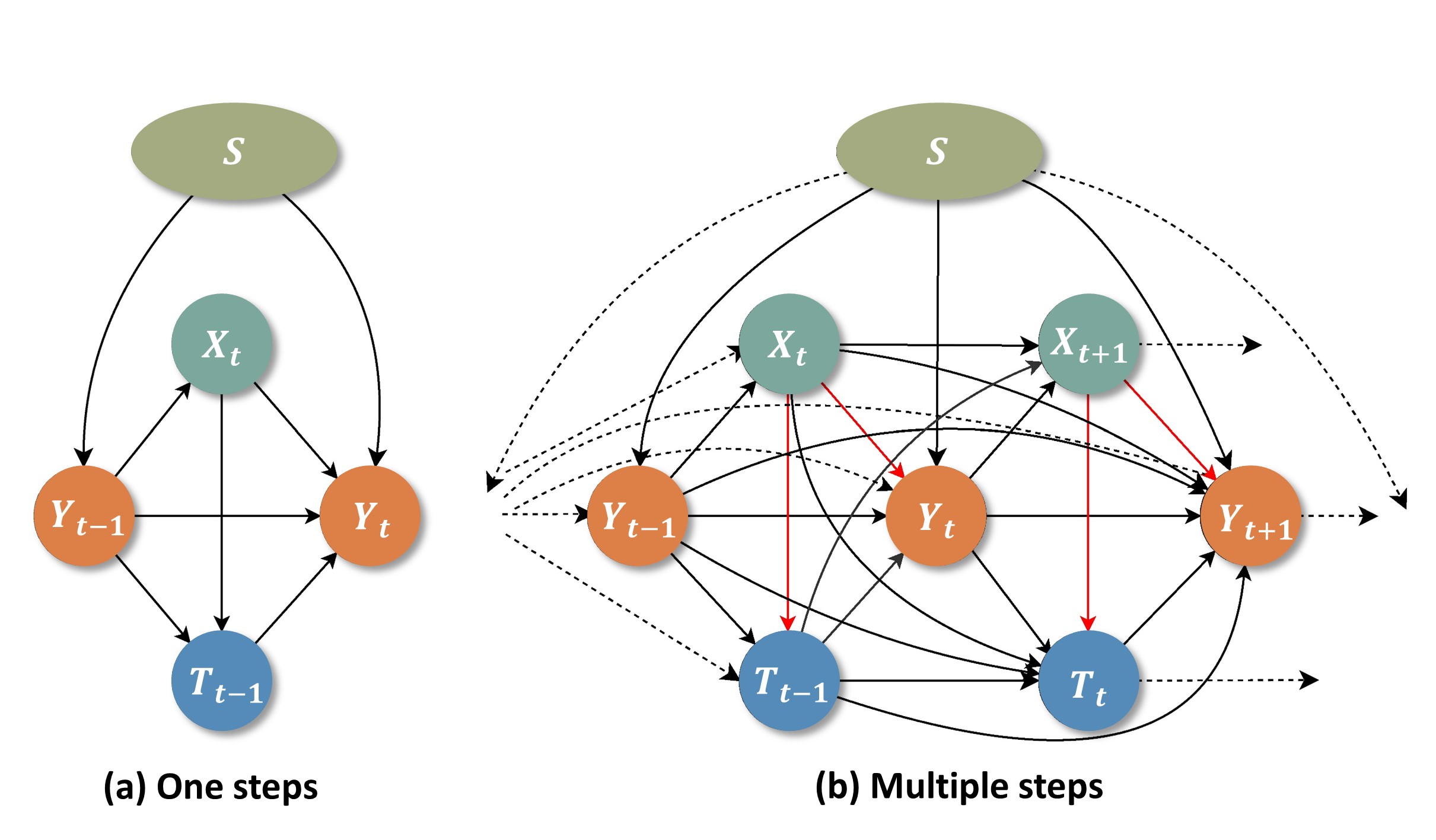}
    \caption{Causal graph for factors interacting mechanism over time.}
    \label{fig_2}
\end{figure}

Fig.\ref{fig_2} illustrates the time-varying causal mechanism of traffic speed under crash treatment. The traffic speed and crashes are effected by historical covariates. Here, we denote ${{\mathbf{\bar H}}_t} = { {{{\mathbf{\bar X}}_t},{{\mathbf{\bar T}}_{t-1}},{{\mathbf{\bar Y}}_t},S}}$ as the history of traffic conditions, where ${{\mathbf{\bar X}}_t} = [{{X_1},\ldots,{X_t}}]$, ${{\mathbf{\bar T}}_{t-1}} = [ {{T_1},\ldots,{T_{t-1}}}]$, and ${{\mathbf{\bar Y}}_t} = [{{Y_1},\ldots,{Y_t}}]$. For each unit, the maximum length of each vector depends on the given time-sequence length of traffic states. Let $\tau\geq1$ denote the prediction horizon for a  $\tau$-step-ahead prediction. Then, we can artificially define the hypothetic intervention vector of treatments as ${{\mathbf{\tilde T}}_{( {t,t + \tau  - 1})}} = [ {{{\tilde T}_t},...,{{\tilde T}_{t + \tau  - 1}}}]$  during $\tau$ time steps and the potential outcomes are ${Y_{t + \tau }}\left( {{{{\mathbf{\tilde T}}}_{\left( {t,t + \tau  - 1} \right)}}} \right)$. Therefore, for a given traffic history and crash intervention, we aim to learn a function $f(\bullet )$  for the expected traffic speed over $\tau$ prediction horizons. The estimated formula can be represented as follows:
\begin{equation}
\setlength{\abovedisplayskip}{3pt}
\setlength{\belowdisplayskip}{3pt}
\label{eq_1}
E({{Y_{t + \tau }}({{{\mathbf{\tilde T}}}_{( {t,t + \tau  - 1})}})|{{{\mathbf{\bar H}}}_t}}) = f({\tau,{{{\mathbf{\tilde T}}}_{({t,t+\tau-1})}},{{{\mathbf{\bar H}}}_t}})
\end{equation}

The learning of function $f( \bullet )$ is a counterfactual task, because we can only observe one type of outcome for each unit at each time step, and then the other unobserved outcomes are referred to as counterfactual outcomes. For example, if we observe a crash at time \textit{t}  for unit \textit{i}, then the traffic speed at next time step is ${Y_{i,t+1}}({{T_{i,t}}=1})$, and it is impossible to have knowledge about ${Y_{i,t+1}}({{T_{i,t}}=0})$. Therefore, when estimating the counterfactual outcomes based on observational data, the confounding bias should be adjusted. Specifically, confounding bias may be raised because the covariates affect outcomes and treatments simultaneously (red arrows in Fig.\ref{fig_2}(b)), which cannot be tackled by traditional machine learning. Additionally, to ensure the potential outcome is identifiable under time-varying treatment settings, we assume three standard assumptions as follows: 

\textbf{Assumption 1.} Sequential consistency: $Y( {{\mathbf{\bar T}}} ) = Y({{{{\mathbf{\bar T}}}^*}})$ if ${\mathbf{\bar T}} = {{\mathbf{\bar T}}^*}$, where ${{\mathbf{\bar T}}^*}$denotes the measured treatment history.  This means that the traffic speed under the given sequence of crash treatments corresponds to the observed speed of units treated by the same crash treatments.

\textbf{Assumption 2.} Sequential positivity: If $P( {{{{\mathbf{\bar H}}}_t} = {{\bar h}_t}} ) \ne 0$, $P( {{T_t}|{{{\mathbf{\bar H}}}_t} = {{\bar h}_t}} )>0$, where ${\bar h_t}$ denotes the factual history of traffic conditions. This suggests that there is a non-zero probability of a crash occurring or not occurring for all instances in the historical data over time.

\textbf{Assumption 3.} Sequential exchangeability: $({{Y_{t +1}}(1),Y_{t+1}(0)}) \bot {T_t}|{{\mathbf{\bar H}}_t}$. This indicates that the likelihood of a crash happening at the current time is independent of potential outcomes, conditioned on the observed history. In other words, there are no hidden confounding factors between treatments and outcomes.

\section{Methods}
\subsection{Marginal Structural Models}
Before predicting the counterfactual traffic speed, we need to briefly introduce the basic methods for time-varying treatment response estimation: Marginal Structural Models (MSMs), which are widely applied for estimating causal effects with IPTW from observational data\cite{robins_marginal_2000},\cite{hernan_marginal_2001}, and the main form can be expressed as follows:
\begin{equation}
\label{eq_2}
E( {{Y_{t + \tau }}|{{{\mathbf{\bar T}}}_{t - 1}},S} ) = {\beta _0} + {{\bm{\beta }}_1}{{\mathbf{\bar T}}_t} + {{\bm{\beta }}_2}S \otimes {{\mathbf{\bar T}}_t} + {{\bm{\beta }}_3}S
\end{equation}
where $\bm{\beta}$ is the vector of parameters to be estimated, in which $\bm{\beta_1}$ are the causal effects of treatments, and $\bm{\beta_2}$ are the causal effect of treatments conditioned on \textit{S}.

Under the three assumptions mentioned above, the unbiased estimates of causal parameters can be obtained by fitting the model with the stabilized weights (SW):

\begin{equation}
\label{eq_3}
{\mathbf{SW}}( {t,\tau } ) = \frac{{\prod\limits_{t = 1}^{t + \tau } {pr( {{T_t}|{{{\mathbf{\bar T}}}_{t - 1}}})} }}{{\prod\limits_{t = 1}^{t + \tau } {pr({{T_t}|{{{\mathbf{\bar H}}}_t}} )} }}
\end{equation}
where $pr(\cdot)$ is the probabilities of treatments conditioned on the covariates.

Commonly, the numerator $pr({{T_t}|{{{\mathbf{\bar T}}}_{t-1}}})$ is referred to as propensity score and the denominator $pr({{T_t}|{{{\mathbf{\bar H}}}_t}})$ is historical propensity score. This method creates a pseudo population by \textbf{SW} to balance the covariates, thus, we could build two networks to fit the propensity score and outcome respectively as in \cite{shi_adapting_2019}.

\begin{figure*}[t]
    \centering
    \setlength{\abovecaptionskip}{0.cm}
    \hspace{-0.3in}
    \includegraphics[ width=7 in]{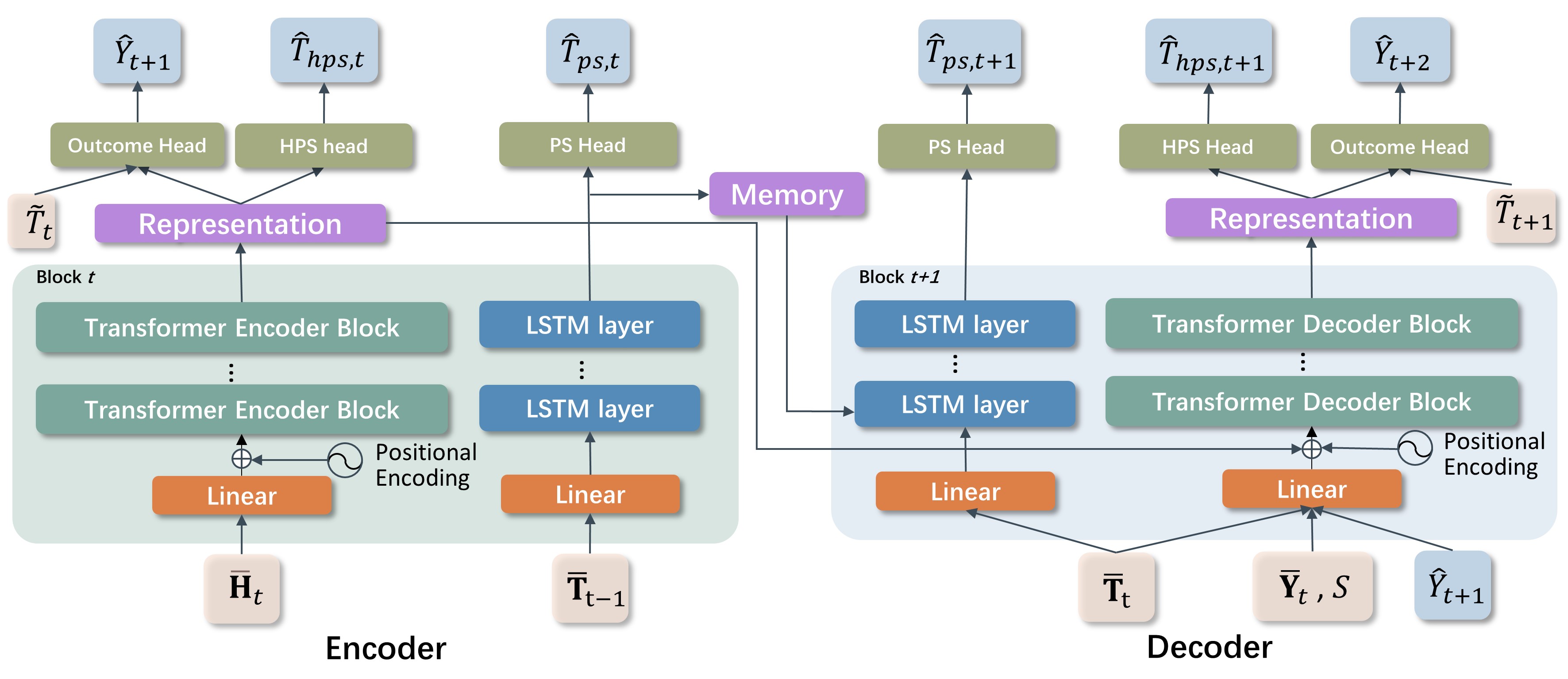}
    \caption{The architecture of MSCT.}
    \label{fig_3}
\end{figure*}

\subsection{Domain Adaptation for Counterfactual Prediction}
Although adjusting by \textbf{SW} can remove bias, the prediction bias increases with the increase in the dimension of covariates. Scholars have found that counterfactual prediction problems can also be transformed into a domain adaptation problems. Robins \cite{robins_association_1999} has proved that when using \textbf{SW}, the bias from time-varying confounders can be removed when $P( {{T_t}|{{{\mathbf{\bar H}}}_t}}) = P( {{T_t}|{{{\mathbf{\bar T}}}_{t - 1}}})$ . Therefore, let $T_t^k$ denote the \textit{k}-th possible treatment (i.e., domain) at time \textit{t}, a representation $\Phi$ of history that could remove the time-varying bias should have the same distribution across the different possible treatments: $P({\Phi({{{{\mathbf{\bar H}}}_t}})|{T_t} = T_t^1})=  \cdots = P({\Phi( {{{{\mathbf{\bar H}}}_t}})|{T_t} = T_t^k})$ \cite{bica_estimating_2020,li_deep_2018}. Because this study only considers crash and non-crash scenarios, there are two domains in this problem.

\subsection{Marginal Structural Causal Transformer}
Inspired by the above methods, we propose a novel transformer model called Marginal Structural Causal Transformer (MSCT) to estimate the counterfactual speed under crash treatment. To predict the traffic speed at multiple time horizons, the model applies sequence-to-sequence architecture, which is depicted in Fig.\ref{fig_3}. The encoders receive all historical features and conduct one-step-ahead prediction. As for decoders, each of them will output the prediction sequentially. For simplicity, this figure only shows the last encoder and the first decoder. The inputs of MSCT include historical traffic statements ${\mathbf{\bar H}}$ and the intervention strategies of treatment $\mathbf{\tilde T}_{t,t+\tau -1}$ (i.e., assumed crashes within a specific time period). When predicting the future horizon, the predictions of outcomes $\mathbf{\hat Y_{({t+1:t +\tau})}}$ are fed to the model autoregressively starting from the predicted time step.

Each MSCT block contains two parallel neural network pathways, one consists of LSTMs and the other consists of Transformers. LSTMs pathway is constructed to predict propensity score ${\hat T_{ps}} = pr( {T|{\mathbf{\bar T}}})$. The Transformers pathway is constructed to predict outcome $\hat Y$  and historical propensity score ${\hat T_{hps}} = pr( {T|{\mathbf{\bar H}}})$. To enhance generalization and reduce computational complexity, we adopt Variational RNN \cite{chung_recurrent_2016}, a regularization technique that applies variational dropout to stochastically set neuron outputs to zero during training. At the last block of the encoder, the outputs of LSTMs will be processed by linear networks and then be input into the decoder as the memory states.

\textbf{Encoder:} The encoder aims to learn a good representation of current traffic states that captures the key information of historical statements. This module will predict one-step-ahead traffic speed \(\hat Y_{t+1}\)  given the observational previous traffic conditions and factual crashes. For time step \textit{t}, the inputs of the encoder are \(\mathbf{\bar H}_t\)  which are firstly concatenated and mapped to a  \(d_h\)-dimension hidden state space \(h_t\) by a fully-connected linear layer. Then, the hidden states for all time steps are input to the first transformer encoder blocks, and the subsequent transformer blocks receive the outputs of the previous blocks as the inputs. Finally, the outputs from the last block are further processed by a fully connected layer followed by an exponential linear unit (ELU) activation function before being taken as a representation. The formalization is as follows:
\begin{align}
    {h_t} &= Linear(Concat( {{\mathbf{\bar X}}_t},{{\mathbf{\bar T}}_{t - 1}},{{\mathbf{\bar Y}}_t},S_t)) \label{eq_4} \\
    {\mathbf{h}}_t^0 &= {[{{h_1}, \ldots ,{h_t}}]^{T}} \label{eq_5} \\
    {\mathbf{h}}_t^l &= TransEncoderBlock_l({{\mathbf{h}}_t^{l-1}}) \label{eq_6} \\
    {\mathbf{\Phi }}_t &= ELU ({Linear({{\mathbf{h}}_t^l})}) \label{eq_7}
\end{align}
where \({\mathbf{h}}_t^l \in {\mathbb{R}^{t \times {d_h}}}\) denotes the hidden states processed by the l-layer block, \(l \in \{ 1,2,..., L\} \), L denotes the total number of blocks, and \({{\mathbf{\Phi }}_t} \in {\mathbb{R}^{t \times {d_h}}} \) is the representation of historical conditions.

\textbf{Decoder:} The following future horizons \({\hat Y_{t + \tau }}\) (\(\tau \geq 2\)) are predicted by the decoder given the representation learned from the encoder. The inputs of the decoder contain the historical treatment \({\mathbf{\bar T}}_{t - 1}\), historical outcomes \({\mathbf{\bar Y}}_t\), static features \textit{S}, and the predict outcomes \({\hat Y_{t + 1}}\) that are autoregressively fed to the model. Different from the encoder, the representation from the encoder is input into the decoder blocks together with the transformed hidden state \(h_t\). Then, to predict the counterfactual outcomes, the intervention strategies \({{\mathbf{\tilde T}}_{({t,t + \tau  - 1} )}}\) and outputs of the last blocks are input into a linear layer to predict the outcomes:
\vspace{-0.2cm}
\begin{align}
    {h_{t + \tau }} &= Linear( {Concat({{T_{t + \tau  - 2}},{Y_{t + \tau  - 2}},S,{{\hat Y}_{t + \tau  - 1}}})}) \label{eq_8} \\
    {\mathbf{h}}_{t + \tau }^0 &= {[{{h_{t + 1}}, \ldots ,{h_{t + \tau }}}]^{T}} \label{eq_9} \\
    {\mathbf{h}}_{t + \tau }^l &= TransDecoderBloc{k_l}( {{\mathbf{h}}_{t + \tau }^{l - 1},{{\mathbf{\Phi }}_t}}) \label{eq_10} \\
    {{\mathbf{\Phi }}_{t + \tau }} &= ELU ( {Linear ( {{\mathbf{h}}_{t + \tau }^l} )}) \label{eq_11}    
\end{align}

\textbf{Output Heads:} There are three heads for outputs, i.e., Outcome Head \({\mathcal{D}_y}({\Phi ({{{{\mathbf{\bar H}}}_t}}),{{\tilde T}_{({t + \tau  - 1})}}})\), Propensity Score Head (PS) \({\mathcal{D}_{ps}}({{{{\mathbf{\bar T}}}_{t-1}}} )\), and Historical Propensity Score Head (HPS) \({\mathcal{D}_{hps}}( {\Phi ( {{{{\mathbf{\bar H}}}_t}})})\). Each head consists of two fully connected linear layers with an ELU activation in between.

\begin{figure}[t]
    \centering
    \setlength{\abovecaptionskip}{0.cm}
    \hspace{-0.3in}
    \includegraphics[ width= 3.5in]{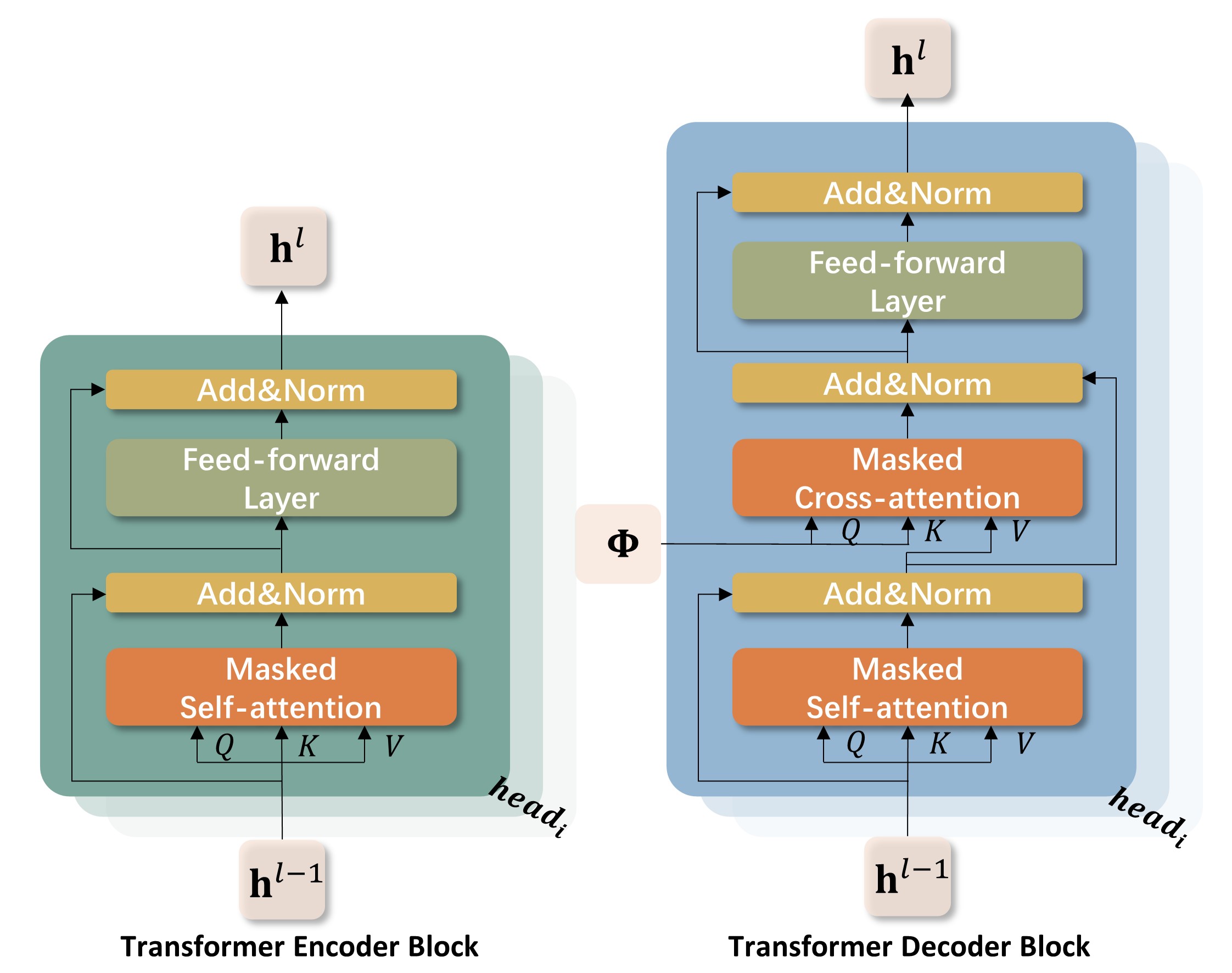}
    \caption{The architecture of transformer blocks.}
    \label{fig_4}
\end{figure}

\textbf{Transformer Block:} Transformers are deep neural networks used for sequential data that commonly employ a specialized attention mechanism \cite{vaswani_attention_2023}. The Transformer architecture is based on the mechanism of self-attention, which allows the model to weigh the importance of different values in a sequence when processing it. The key components of a Transformer model include multi-head attention, position-wise feed-forward layer, and Add\&Norm layer (Shown as Fig. \ref{fig_4}).
(1) Multi-head attention is a module that runs through an attention head several times in parallel. An attention mechanism can be described as mapping queries \textit{Q}, keys \textit{K}, and values \textit{V} to an output. The hidden states \(\mathbf{h^l}\) are processed by three linear layers to \(d_a\)-dimension \textit{Q}, \textit{K}, and \textit{V}, respectively. As designed in  \cite{vaswani_attention_2023} (called scaled dot-product attention), we compute \textit{i}-th head of output as follows:
\begin{equation}
    \setlength{\abovedisplayskip}{3pt}
    \setlength{\belowdisplayskip}{3pt}
    \label{eq_12}
    Attentio{n^i}({{Q^i},{K^i},{V^i}}) = \operatorname{softmax}({\frac{{{Q^i}{K^i}^{\top }}}{{\sqrt {{d_a}} }}})V
\end{equation}

Then, the multi-head attention processes concatenated heads by linearly projection:
\begin{equation}
    \label{eq_13}
    MultiHead({Q,K,V})=Concat({head_1},...,{head_i}){W^O}
\end{equation}
\begin{equation}
    \label{eq_14}
    head_i = Attention^i({Q^i}W_i^Q,{K^i}W_i^K,{V^i}W_i^V)
\end{equation}
where \(W_i^Q \in \mathbb R^{d_h \times d_a}\), \(W_i^K \in \mathbb R^{d_h \times d_a}\),\(W_i^V \in \mathbb R^{d_h \times d_a}\), and \(W^O \in {\mathbb R^{{i}{d_h} \times d_a}}\) are parameters for training.

In addition, multi-head attention is used in the encoder and decoder by different ways. As shown in Fig. 4, the encoder only contains self-attention which only takes the hidden states from the same subnetwork as inputs. The decoder contains an extra cross-attention layer that receives the queries from the previous decoder block and keys and values from the output of the encoder as inputs. Therefore, self-attention and cross-attention can be formalized as follows:
\begin{align}
    \label{eq_15}
    Self(Q,K,V) &= [{Q,K,V}] \nonumber \\ 
   &= [Linear({\mathbf{h}}^{l - 1}),Linear( {\mathbf{h}}^{l - 1}),Linear( {\mathbf{h}}^{l - 1})] \\
   Cross(Q,K,V) &= [{Q,K,V}] \nonumber \\ 
   &= [Linear({\mathbf{h}}^{l - 1}),Linear( {\mathbf{\Phi}}),Linear( {\mathbf{\phi}})]
\end{align}

To prevent future information flows in the current state, mask matrixes are used in scaled dot-product attention to set the future input values of softmax to \(- \infty \).

(2) A Position-wise feed-forward layer is applied after each attention and takes the outputs of multi-head attention as inputs. This layer consists of two fully connected linear layers with a ReLU activation in between:
\begin{equation}
    \label{eq_17}
    FFL(\mathbf{\alpha}) = Linear(ReLU(Linear(\mathbf{\alpha})))
\end{equation}
where \(\mathbf{\alpha}\) denotes the outputs of the sub-layer in the blocks.
(3) Add\&Norm layer functions as the connection between each layer (attention, feed-forward) in all encoder and decoder blocks. It has a residual connection and is followed by a layer-normalization step \cite{ba_layer_2016}. The computation can be formalized as follows:
\begin{equation}
    \setlength{\abovedisplayskip}{3pt}
    \setlength{\belowdisplayskip}{3pt}
    \label{eq_18}
    \mathbf{\alpha} = LayerNorm(\mathbf{\alpha} + Sublayer(\mathbf{\alpha}))
\end{equation}

\textbf{Positional Encoding:} To provide the sequence information of hidden states for the transformer, absolute positional encoding is utilized to the linear-transformed inputs right before the first transformer block \cite{vaswani_attention_2023}. The fixed weights are applied to encode each time step \textit{t} by sine and cosine function:
\begin{align}
    \label{eq_19}
    PE_{(t,2d)} &= sin(t/{1000^{2d/{d_h}}})  \\ 
    PE_{(t,2d+1)} &= cos(t/{1000^{2d/{d_h}}})
\end{align}
where \textit{d} is the dimension. Thus, each dimension of the positional encoding corresponds to a sinusoid, enabling linear transformations to be performed between neighboring time steps and time-delta shifts.

\subsection{Training Procedure}
The key procedure for counterfactual prediction is removing the time-varying confounding bias by training. The aim of our models is: (a) predict the outcome by representations, (b) balance the historical treatments by fitting propensity score, and (c) remove the correlation between representations and treatments.
For objective (a), the parameters of networks for outcomes and representations can be fitted by the true next outcome via minimizing the mean squared loss (MSE):
\begin{equation}
    \label{eq_21}
    \mathcal{L}_{t,y}({{\theta _Y},{\theta _R}}) = {\|Y_{t + 1}-{\mathcal{D}_y}({{\Phi _t}({{{{\mathbf{\bar H}}}_t};{\theta _R}});{\theta _Y}})}\|^2
\end{equation}

For objective (b), we need to fit the propensity score network according to the next true treatment by binary cross entropy (BCE) loss:
\begin{equation}
    \label{eq_22}
    \mathcal{L}{_{t,ps}}( {{\theta _T},{\theta _{PS}}}) =  - \sum\limits_{k = 1}^K {{\mathcal{I}_{\{{T_t} = T_t^k\}}}\log {\mathcal{D}_{ps}}( L( {{{{\mathbf{\bar T}}}_{t - 1}};{\theta _T}} );{\theta _{PS}})}
\end{equation}
where \(\mathcal{I}\) is the indicator function, and \(L(\bullet)\) represents the LSTMs pathway in MSCT blocks.

Objective (c) requires learning a representation that cannot be used to predict treatment, thus, a domain generalization loss function is applied \cite{tzeng_simultaneous_2015}: 
\begin{equation}
    \label{eq_23}
    \mathcal{L}{_{t,hps}}({{\theta _R},{\theta _{HPS}}}) =  - \sum\limits_{k = 1}^K {\frac{1}{K}} \log {\mathcal{D}_{hps}}( \Phi _t ( {{{{\mathbf{\bar H}}}_{t}};{\theta _R}} );{\theta _{HPS}})
\end{equation}

By combining all the above loss functions, the total loss can be obtained as follows:

\begin{align}
    \label{eq_24}
  &{R_t}( {{\theta _R},{\theta _Y},{\theta _T},{\theta _{PS}},{\theta _{HPS}}} ) \nonumber \\
  &= \mathcal{L}{_{t,Y}}( {{\theta _R},{\theta _Y}}) + \lambda ( {\mathcal{L}{_{t,ps}}( {{\theta _T},{\theta _{PS}}} ) + \mathcal{L}{_{t,hps}}( {{\theta _R},{\theta _{HPS}}} )})
\end{align}
where \(\lambda\) is the hyperparameter controlling the trade-off between domain confusion and outcome prediction.

We realized MSCT by PyTorch Lightning and used Adam to optimize (\(\Delta\) ) the loss with learning rate \(\eta\) \cite{kingma_adam_2017}. The overall training procedure consists of three steps, and we describe the training process by pseudocode in \textbf{Algorithm \ref{alg_1：AOA}}.
\begin{minipage}{.5\textwidth}
\setlength{\footnotesep}{0.2pt}
\begin{algorithm}[H]
    \caption{ Training Procedure}
    \label{alg_1：AOA}
    \renewcommand{\algorithmicrequire}{\textbf{Input:}}
    \renewcommand{\algorithmicensure}{\textbf{Output:}}
    \begin{algorithmic}[]
        \REQUIRE Initialize dataset \({{\mathbf{\bar H}}_t} = \{ {{{\mathbf{\bar X}}_t},{{\mathbf{\bar T}}_{t - 1}},{{\mathbf{\bar Y}}_t},S} \}\), intervention \({{\mathbf{\tilde T}}_{({t,t + \tau  - 1} )}}\), and hyperparameters \footnote{Hyperparameters in this model include \textit{number of epochs}, \textit{batch size}, \textit{drop rate}, \textit{size of hidden states},\textit{ number of heads}, and\textit{ number of layers}.}.
        \ENSURE Optimized parameters: \({{\mathbf{\Theta }}_E},{\mathbf{\Theta }}_D\)
    \end{algorithmic}
    \textbf{Step 1:} Training Encoder: Optimize parameters \\
    \({{\mathbf{\Theta }}_E} = {\left( {{\theta _R},{\theta _Y},{\theta _T},{\theta _{PS}},{\theta _{HPS}}} \right)_E}\)
    \begin{algorithmic}[]    
    \FOR{ each epoch}
        \FOR{ batch B in dataset}
        \STATE \({( {{\theta _R},{\theta _T},{\theta _{PS}}})_E} \leftarrow {( {{\theta _R},{\theta _T},{\theta _{PS}}} )_E} - \eta \Delta ( {\frac{1}{B}\sum\nolimits_B \!{\sum\nolimits_t \!{( {\mathcal{L}{_{t,y}}\! + \lambda \mathcal{L}{_{t,ps}}} )} } })\)
            \STATE \({( {{\theta _R},{\theta _Y},{\theta _{HPS}}})_E} \leftarrow {( {{\theta _R},{\theta _Y},{\theta _{HPS}}} )_E} - \eta \Delta ( {\frac{1}{B}\sum\nolimits_B {\sum\nolimits_t {( {\mathcal{L}{_{t,y}} + \lambda \mathcal{L}{_{t,hps}}} )} } })\)
        \ENDFOR
    \ENDFOR
    \RETURN Encoder network: \(Encoder({\mathbf{\bar H}}_{i,t};{\mathbf{\Theta}}_E)\)
    \end{algorithmic}
    \textbf{Step 2:} Compute the encoder representations and memory states \( {\mathbf{\Phi},\mathbf M}\)
    \begin{algorithmic} []
    \FOR{ unit \textit{i}=1 to \textit{N}}
        \FOR{ \textit{t}=1 to \textit{T}}
            \STATE \(( {{\Phi _{i,t}},{M_{i,t}},{{\hat Y}_{i,t + 1}}} ) = Encoder({{\mathbf{\bar H}}_{i,t}};{{\mathbf{\Theta }}_E})\)
        \ENDFOR
    \ENDFOR
    \end{algorithmic}
    \textbf{Step 3:} Training Decoder: Optimize parameters \({{\mathbf{\Theta }}_D} = {\left( {{\theta _R},{\theta _Y},{\theta _T},{\theta _{PS}},{\theta _{HPS}}} \right)_D}\)
    \begin{algorithmic}[]
    \FOR{ epochs }
        \FOR{ batch B in dataset \(\{ {{{\mathbf{T}}_{t:t + {\tau _{\max }} - 1}},{{\mathbf{Y}}_{t + 2:t + {\tau _{\max }}}},S} \} \cup \{ {{{\mathbf{\Phi }}_t},{{\mathbf{ M}}_t},{{{\mathbf{\hat Y}}}_{t + 1}}} \}\)}
            \STATE  \({( {{\theta _R},{\theta _T},{\theta _{PS}}})_D} \leftarrow {( {{\theta _R},{\theta _T},{\theta _{PS}}} )_D} - \eta \Delta ( {\frac{1}{B}\sum\nolimits_B {\sum\nolimits_{t = 2}^{{\tau _{\max }}} {( {\mathcal{L}{_{t,y}} + \lambda \mathcal{L}{_{t,ps}}})} } })\)
            \STATE \({( {{\theta _R},{\theta _Y},{\theta _{HPS}}})_D} \leftarrow {( {{\theta _R},{\theta _Y},{\theta _{HPS}}} )_D} - \eta \Delta ( {\frac{1}{B}\sum\nolimits_B {\sum\nolimits_{t = 2}^{{\tau _{\max }}} {( {\mathcal{L}{_{t,y}} + \lambda \mathcal{L}{_{t,hps}}} )} } })\)
        \ENDFOR
    \ENDFOR      
    \RETURN Encoder network: \(Decoder({\mathbf{\bar H}}_{i,t};{\mathbf{\Theta}}_D)\)
    \end{algorithmic}
\end{algorithm}
\vspace{-\baselineskip}
\end{minipage}

\section{EXPERIMENTS}
Counterfactual outcome prediction is different from traditional prediction tasks because of the unobserved potential outcomes, which means there is no ground truth for real-world data. Therefore, we propose a way to generate a synthetic dataset based on common sense and the causal mechanism of traffic features. We also use a real-world dataset about highway traffic crashes. Due to no access to the true counterfactuals, we only present the performance of factual outcomes which can also prove the efficiency of our model to some degree.

\subsection{Synthetic Data Generation}
In this section, A DGP is proposed for traffic speed considering the impact of crashes. Here, we hold the belief that traffic speed \textit{Y} is influenced by crashes \textit{T} and some other factors (represented by \textit{X}), and the occurrence of crashes is also affected by these factors. Therefore, the procedure of synthetic data generation considers the basic traffic temporal tendency (i.e., the peak and non-peak hours), the temporal dependency, the dissipation of congestion, and the time-varying confounding effects. For each unit, we generate data iteratively as follows:
\begin{align}
    \label{eq_25}
      Y( t) &= ( {{\beta _1}{X_t} - {\beta _2}{T_t} + \varepsilon }) \times Y( {t - 1}) \nonumber \\ 
       &+ Base( t ) \times {({Y( {t - 1} )}/{Base({t - 1})})^{g(d)}}
\end{align}

\begin{figure*}[hb]
    \centering
    \includegraphics[ width=7 in]{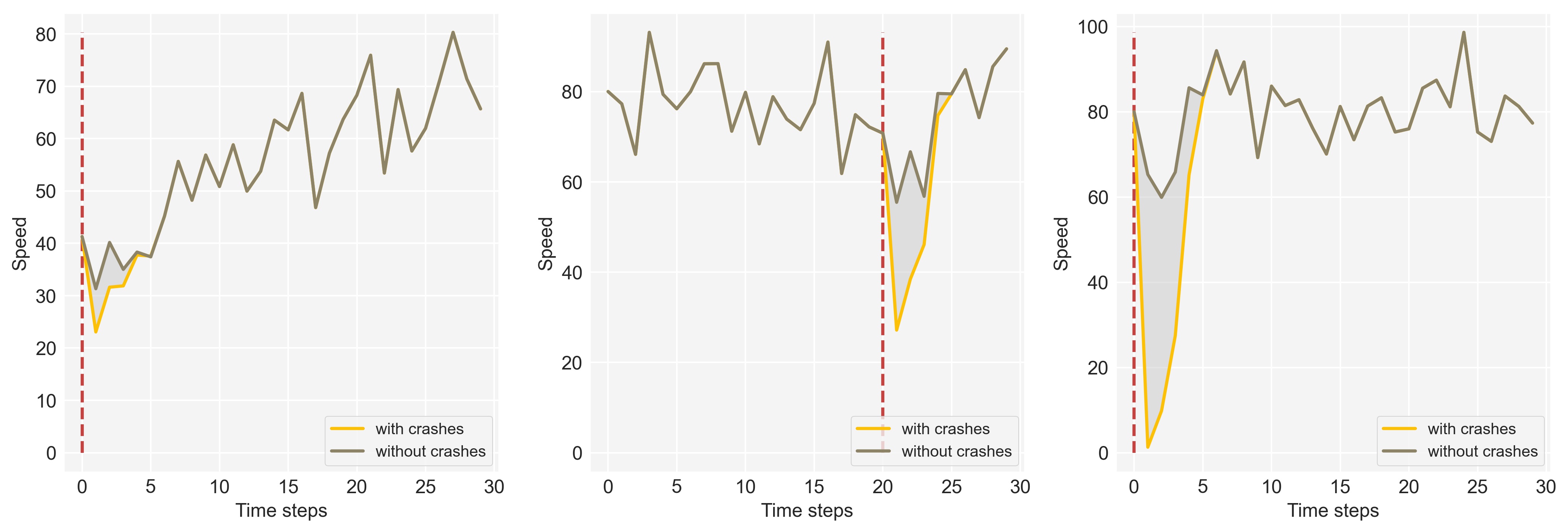}
    \caption{Synthetic traffic speed data samples. Red dashed line denotes the time of crash occurrence.}
    \label{fig_5}
\end{figure*}

This formula can be divided into three parts:

1) The first part \(( {{\beta _1}{X_t} - {\beta _2}{T_t} + \varepsilon }) \times Y( {t - 1} )\) simulates the time-varying causal effects of treatments and covariates. \({X_t} \sim N({0,1} )\) is the confounder for treatment and outcome. Here, we use a one-dimension covariate for confounding and assume its coefficient \(\beta _1 = 0.1\).  \(T_t\) is the treatment indicator for time step \textit{t} ( \(T_t=1\) indicates a crash occurs, while \(T_t=0\) represents no crash), and it is affected by \(X_t\) . Therefore, we generate the treatment by the following formalization:
\begin{align}
    \label{eq_26}
    T_t=\begin{cases}
      1& \text{ if } \overleftarrow{X}_t \ge  \overleftarrow{x}_{10th}\\
      0& \text{ if } \overleftarrow{X}_t <   \overleftarrow{x}_{10th}
    \end{cases} \\
    \overleftarrow{X}_{t,\omega} = \frac{1}{\min (t,\omega)}\sum_{\max (0,t-\omega)}^{t} X_t  
\end{align}
where \(\overleftarrow{X}_{t,\omega}\) denotes the average of historical  -time-step covariates, which reflects the \textbf{degree of time-varying confounding bias}. \(\overleftarrow{x}_{10th}\)  denotes the 10-th percentile value of  \(\overleftarrow{X}_t\)  for all sample points.

Considering the varying levels of crash impact, the coefficients \(\beta_2\)  are determined based on the following assumptions: there are three types of crashes and their respective causal effects are \( \beta_2 = [0.2,0.4,0.8]\). For each type of crash, the probability of occurrence is \( p_c=[0.6,0.3,0.1]\). By this setting, the probability of the most serious crash occurring is 10\%.

 2) The second part is \(Base(t)\), which simulates the basic trends of daily changes in traffic speed based on the normal distribution. We assume the formalization of basic trends is:
\begin{equation}
    Base(t)= \Psi-\lambda \times \frac{exp(-(\phi(t)-\mu)^2/{2\sigma^2}}{\sigma\sqrt{2\pi}}
\end{equation}
where \(\Psi\) is the assumed maximum basic speed,  \(\lambda\) is the amplitude,  \(\mu\) is the mean, and \(\sigma\) is the variance. These parameters can control the change trends. Here, we set  \(\Psi=80\), \(\mu=6\), and \(\sigma=1\) .Thus, \(\phi(t) \in \{1,2,\dots,12\}\). Because normal distribution is an unimodal curve that can simulate one peak for traffic conditions, we apply this function to simulate the traffic speed trends for half a day. To generate the data by 5-min interval for a whole day (i.e., 720 values of time for one day), we can set \(\phi(t)={t \% 360 }/30\), where \(t \in \{1,2,\dots,720\}\).

3) The third part \(({Y( {t - 1} )}/{Base({t - 1})})^{g(d)}\) reflects the dissipation of crash influence. When the speed at time \textit{t-1} is affected by crashes, then the speed at time step \textit{t} will also be influenced, but to a lesser extent. Therefore, the function \(g(d)\) represents the impact duration of crashes. This study assumes that the crash will affect the next 5 time steps, then:
\begin{equation}
    g(d)=\begin{cases}
        1.25-0.25d& \text{if } 1\leq d=(t-t_c) \leq 5 \\
        0 & \text{others}
    \end{cases}
\end{equation}
where \(t_c\) denotes the time when the crash occurs. To illustrate the synthetic traffic speed data affected by crashes, we generate three samples with 30-time-steps sequential data shown in Fig. \ref{fig_5}.

To test the performance of counterfactual prediction, for each unit in the test set and each time step, we simulate several counterfactual speeds under the assumed intervention strategies \(\mathbf{{\tilde T}}_{(t,t+\tau-1)}\). For one-step-ahead prediction, we simulate both crash and non-crash speed \(Y_{t+1}\). For multi-step-ahead prediction, a “single sliding treatment” setting is applied where strategies involve only a single treatment but iteratively move it over a window ranging from \textit{t} to \(t+\tau_{max}-1\) \cite{bica_estimating_2020}. In addition, we also consider non-crash strategies (i.e., full zero vector). Therefore, the number of all potential outcomes grows exponentially. For example, when \(\tau_{max}=2\), for each time step in each unit, we will simulate three counterfactual outcomes(i.e., \(\mathbf{{\tilde T}}_{(t,t+1)}=\{[1,0],[0,1],[0,0]\}\)).

It should be noted that, during the training procedure, only one type of potential outcome (i.e. observed outcomes) can be input into the model. During the test procedure, all generated counterfactual outcomes for each unit should be input to test the performance. This ensures that the model has learned the ability to predict counterfactual outcomes from observational data. 

\subsection{Real-world Data}
The real-world data used in this study was collected from the main road of Interstate 5 (I5) in Washington from Nov. 2019 to May. 2021, and the milepost range from 139 to 178. The traffic data consists of traffic speed, occupancy, and volume, and the incident data set collected by the Highway Safety Information System provides information on crashes including time, location, crash type, etc. We matched all data according to the time, location, and direction and aggregated them in one-mile and 5-minute intervals.

We use average speed as the outcome and the indicator of crash occurrence as the treatment for each time step. Due to the varying effects of different types of crashes, the value of treatment is 0 to 3, corresponding to four scenarios: no crash, crash to objects (OBJ), sideswipe (WIPE), and rear-end (REAR). This treatment variable is then encoded into a one-hot format when fed into the model. The time-varying confounders including \textit{time}, \textit{the max speed difference between lanes}, \textit{occupancy}, \textit{volume}, and \textit{congestion index} (real speed divided by speed limit). Moreover, considering the impacts of other locations, we also incorporate the average speed and congestion index at one and two miles upstream and downstream of the crash location. The static features are \textit{milepost}, \textit{direction}, \textit{day of week}, and \textit{weather}. Because the traffic data are continuous long time series, we utilize a sliding window to generate the data sequence from the traffic flow data. 

\subsection{Baseline Models and Evaluation Metrics}
To compare the performance of counterfactual prediction between causal-based and non-causal-based models, we first selected four traditional temporal deep learning models, namely RNN, LSTM, Bi-LSTM, and GRU, to serve as baseline models. In addition, we also chose several causal models in the state-of-the-art literature.These are:  RMSNs \cite{lim_forecasting_2018}, CRN \cite{bica_estimating_2020}, G-Net \cite{li_g-net_2021}, and CT \cite{melnychuk_causal_2022}. All these models are realized by Pytorch Lightning (version 1.4.5) on a GPU-equipped workstation, and their hyper-parameters are tuned by Ray tune. 

To evaluate the performance of the trained model, the Root Mean Square Error (RMSE) is calculated for each test dataset. In addition, because we know the true factual and counterfactual speed for synthetic data, we can calculate the true and predicted causal effects for each crash, and the RMSE for causal effects (CRMSE) can be calculated to evaluate the performance. For the   time step after crash, the calculation of CRMSE can be formulated as follows:
\begin{small}
\begin{align}
    {C\!R\!M\!S\!E}_t &= \sqrt{\frac{1}{N}\! \sum_{i=1}^{N}(c_{\tau}\!-\!\hat{c}_\tau)^2} \nonumber \\
    &= \sqrt{\frac{1}{N} \!\sum_{i=1}^{N}((Y_{true,\tau}^1\!-\!Y_{true,\tau}^0))\!-\!(Y_{pred,\tau}^1\!-Y_{pred,\tau}^0))^2}
\end{align}
\end{small}

Where \(c_{\tau}\) and \(\hat{c}_\tau\) denote the true and predicted crash causal effects, respectively.  \(Y^1\) and \(Y^0\) denote the traffic speed under conditions of a crash occurring and not, respectively.

\section{results}
\subsection{Results of synthetic data}
We generated 1000 samples to train our model, 100 samples for validation, and 100 for testing. For each sample, the length of the sequence is 60 time steps, and \(\tau_{max}\) for the single sliding treatment setting is five. Therefore, we conducted a 6-step-ahead prediction of counterfactual outcomes (the first step is predicted by the encoder, and the others are predicted by the decoder). It should be noted that each sample in the validation and testing sets was extended to include its counterfactual outcomes for evaluating the model's performance. Specifically, during the one-step-ahead prediction conducted by the encoder, we generated non-crash data for each crashed sample. For the multi-step-ahead prediction conducted by the decoder, when predicting the target speed from time \textit{t+1} to time \textit{t+5}, the treatment indicators from time \textit{t} to time \textit{t+4} were manipulated according to the single sliding treatment settings. Thus, for each current time step, we generated six different future outcomes for every sample. The results are presented in TABLE \ref{tab_1}. According to the findings, our model outperformed other models, especially as the prediction horizon increased, suggesting that our model demonstrates superior performance in longer forecasting horizons.

\begin{table}[!ht]
    \renewcommand{\arraystretch}{1.2}
    \centering
    \caption{RMSE RESULTS FOR SYNTHETIC TRAFFIC DATA (\(\omega=5\))}
    \label{tab_1}
    \begin{tabular}{c|c:c c c c c}
    \hline
        ~ & \(\tau=1\) & \(\tau=2\) & \(\tau=3\) & \(\tau=4\) & \(\tau=5\) & \(\tau=6\) \\ \hline
        LSTM  & 8.92 & 12.61 & 16.88 & 20.45 & 24.59 & 28.07  \\ 
        BiLSTM  & 8.66 & 10.88 & 12.92 & 14.15 & 15.26 & 16.38  \\ 
        GRU  & 9.38 & 11.43 & 13.49 & 14.77 & 15.72 & 22.14  \\ 
        RNN  & 8.73 & 10.57 & 12.61 & 13.87 & 14.55 & 14.75  \\ 
        \hdashline[0.8pt/2.5pt]
        MSMs  & 10.14 & 12.47 & 14.86 & 16.81 & 18.12 & 18.92  \\ 
        RMSN  & 8.27 & 11.58 & 12.83 & 13.10 & 12.85 & 12.78  \\ 
        CRN  & 8.28 & 11.63 & 12.5 & 12.93 & 13.29 & 13.60  \\ 
        G-Net  & 8.31 & 11.08 & 13.2 & 14.07 & 14.54 & 14.68  \\ 
        CT  & 8.09 & 10.98 & 12.84 & 12.65 & 12.78 & 13.02  \\ \hline
        MSCT  & \textbf{7.96} & \textbf{10.47} & \textbf{11.77} & \textbf{11.99} & \textbf{11.86} & \textbf{11.86}  \\ \hline
    \end{tabular}
\end{table}

\begin{figure*}[ht]
    \centering
    \setlength{\abovecaptionskip}{0.cm}
    \includegraphics[ width=7 in]{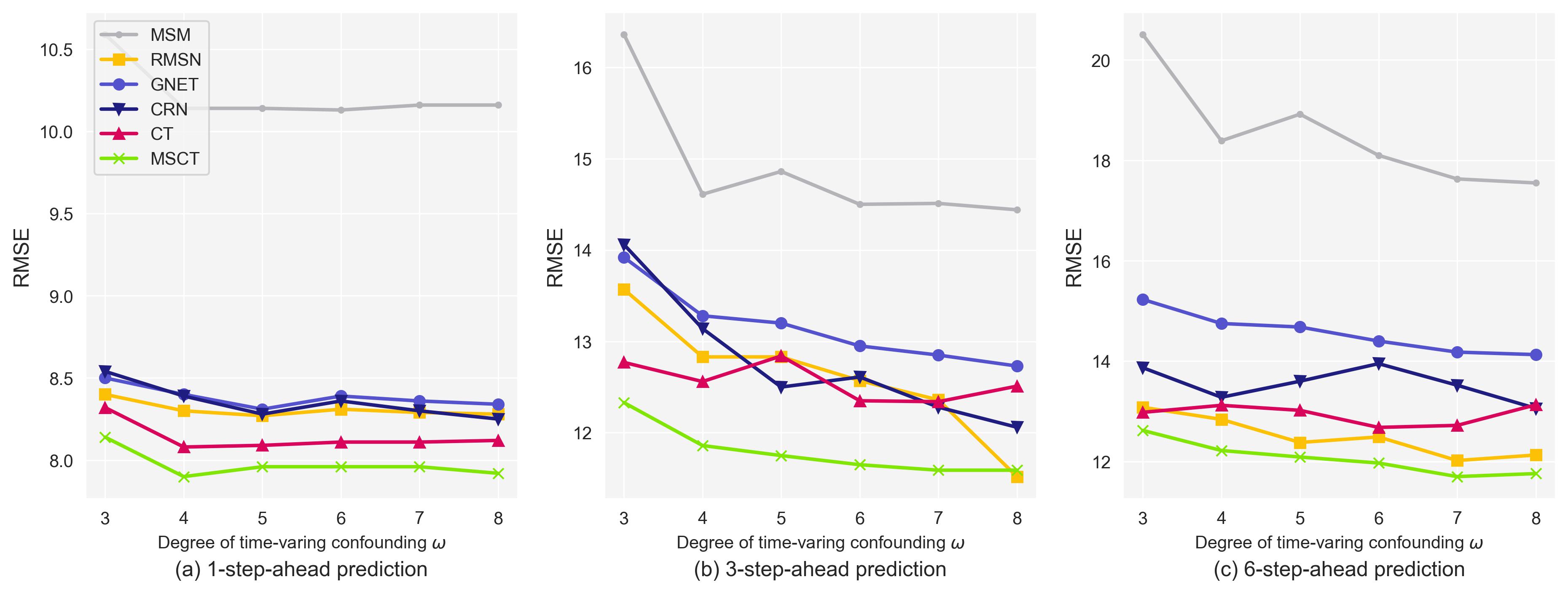}
    \caption{The impact of different degrees of time-varying confounding on RMSE}
    \label{fig_6}
\end{figure*}

Fig. \ref{fig_6}. illustrates the variation in prediction results for different time-varying confounding coefficients \(\omega\). The overall trend of the prediction error is decreasing as  \(\omega\) increases. This phenomenon can be attributed to the fact that when the occurrence of a crash is influenced by a longer historical period, its outcome becomes more predictable, allowing for a better estimation of counterfactual outcomes. For instance, if a crash is caused by prolonged congestion, both the occurrence of the crash and the post-crash traffic conditions are more likely to be anticipated. In contrast, unexpected crashes may be influenced by short-term factors, making their outcomes more challenging to predict. 

The above errors are calculated by all time-step speeds including crash and non-crash conditions. To test the ability of models for causal effects estimation, we further focus on the time steps after the crash occurred. The CRMSE for causal effects prediction is outlined in TABLE \ref{tab_2}. Because we set \(\tau_{max}=5\) in this study, we can only calculate the causal effects for up to five time steps after the crash. According to the results, causal-based deep learning methods exhibit superior performance compared to traditional temporal deep learning models. In addition, MSCT consistently performs well across nearly all-time steps, showing lower CRMSE values compared to other models, indicating better capability of heterogeneous causal effects estimation.

\begin{table}[!h]
    \renewcommand{\arraystretch}{1.2}
    \centering
    \caption{CRMSE RESULTS FOR SYNTHETIC TRAFFIC DATA (\(\omega=5\))}
    \begin{tabular}{c|c c c c c}
    \hline
        ~ & \(\tau=1\) & \(\tau=2\) & \(\tau=3\) & \(\tau=4\) & \(\tau=5\) \\ \hline 
        LSTM  & 15.33 & 14.95 & 12.67 & 12.22 & 10.49  \\ 
        BiLSTM  & 15.81 & 17.34 & 16.74 & 19.00 & 25.17   \\ 
        GRU  & 17.31 & 16.60 & 12.03 & 10.87 & 8.49   \\ 
        RNN  & 14.74 & 15.49 & 13.51 & 13.18 & 15.12   \\ 
        \hdashline[0.8pt/2.5pt]
        MSMs & 16.57 & 15.01 & 12.4 & 11.15 & 10.11  \\
        RMSN & 14.84 & 17.26 & 9.29 & 6.72 & 4.08  \\ 
        CRN & 14.58 & 15.32 & 8.26 & 5.48 & 4.38  \\ 
        G-Net & 16.17 & 18.79 & 13.96 & 11.26 & 10.35  \\ 
        CT & 16.01 & 19.48 & 19.48 & 10.81 & \textbf{3.62}  \\ \hline
        MSCT & \textbf{13.62} & \textbf{14.70} & \textbf{8.05} & \textbf{5.38} & 4.67   \\ \hline
    \end{tabular}
    \label{tab_2}
\end{table}

\begin{figure}[!hb]
    \centering
    \includegraphics[ width= 3.5in]{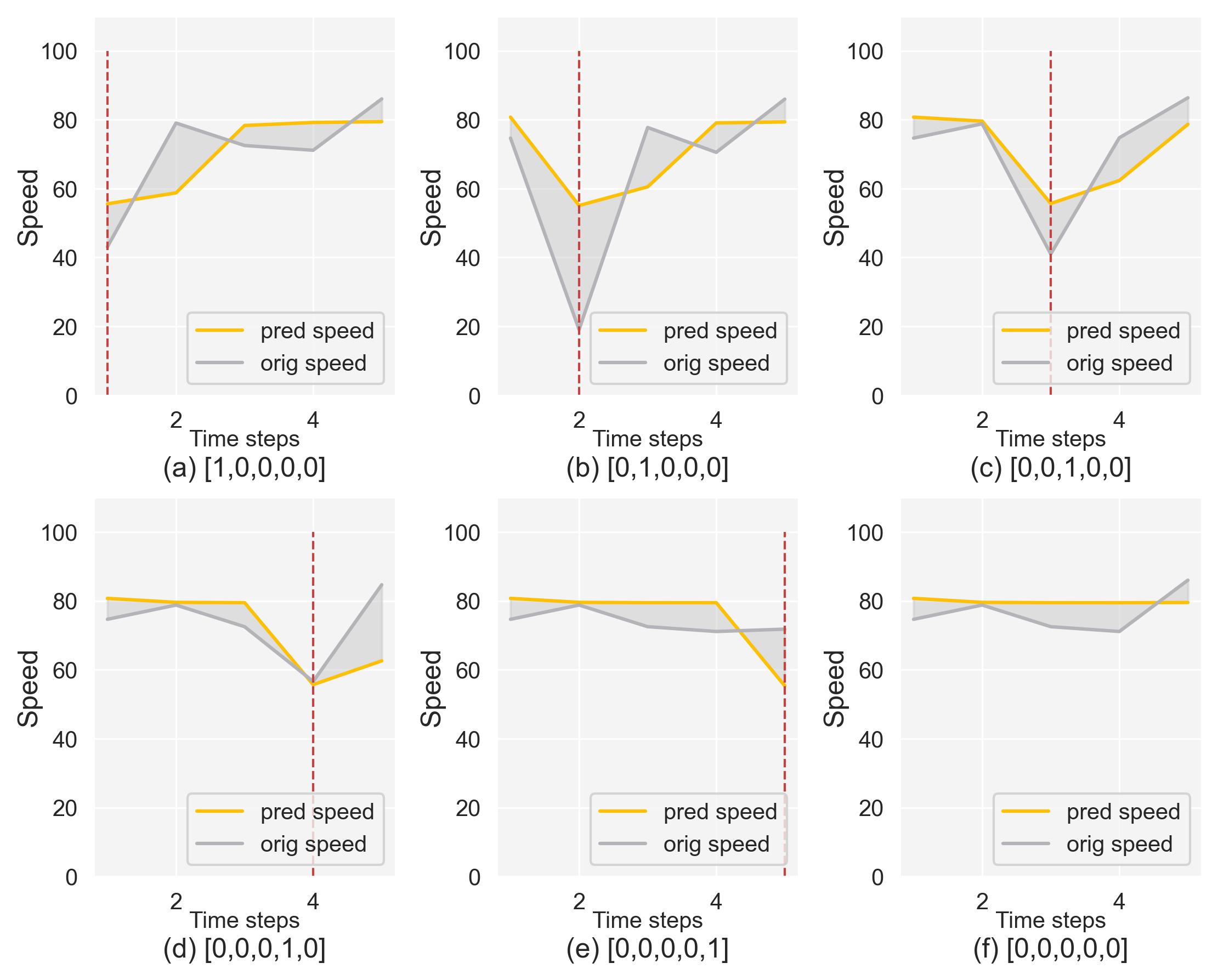}
    \caption{Prediction performance of MSCT for different treatment strategies. Red dashed line denotes the first time step after crashes.}
    \label{fig_7}
\end{figure}

Fig.\ref{fig_7} shows the prediction results of MSCT for a unit under six intervention strategies, with Fig.\ref{fig_7}(a)-(e) under crash conditions, and no crash occurs in Fig.\ref{fig_7}(f). According to the figure, a noticeable decline is easily observed following crashes, and speeds gradually recover over time. This further illustrates the crash-aware capability of the proposed model.

\subsection{ Results of real-world data}

TABLE \ref{tab_3} demonstrates the prediction performance based on real-world data. We randomly selected 3000 samples from real-world datasets, with 30\% of the samples containing crashes. Each sample is a time series with 60 lengths, representing 300 minutes. As we do not have access to actual counterfactual data, all the results are validated based on real-world data only. The results consistently show that our model exhibits better performance compared to other models in overall predictions.

\begin{table}[!h]
    \renewcommand{\arraystretch}{1.2}
    \centering
    \caption{RMSE RESULTS FOR REAL-WORLD TRAFFIC DATA (CRASH RATIO = 0.3)}
    \label{tab_3}
    \begin{tabular}{c|c:c c c c c}
    \hline
        ~ & \(\tau=1\) & \(\tau=2\) & \(\tau=3\) & \(\tau=4\) & \(\tau=5\) & \(\tau=6\) \\ \hline
        LSTM  & 0.92 & 8.37 & 8.71 & 8.63 & 8.70 & 8.83  \\ 
        BiLSTM  & 1.11 & 8.03 & 8.56 & 8.70 & 8.74 & 8.76  \\ 
        GRU  & 1.01 & 7.63 & 8.34 & 8.57 & 8.72 & 8.83  \\ 
        RNN  & 2.64 & 5.15 & 6.26 & 7.06 & 7.48 & 7.32  \\ \hdashline[0.8pt/2.5pt]
        MSMs & 1.35 & 2.98 & 4.40 & 5.88 & 7.52 & 9.20  \\ 
        RMSN & 1.17 & 2.87 & 3.64 & 4.2 & 4.74 & 5.32  \\ 
        CRN & 0.98 & 2.92 & 3.84 & 4.57 & 5.13 & 5.58  \\
        G-Net & 3.4 & 4.23 & 4.85 & 5.42 & 5.97 & 6.47  \\ 
        CT & 0.95 & 2.74 & 3.66 & 4.28 & 4.76 & 5.16  \\ \hline
        MSCT & \textbf{0.83} & \textbf{2.68} & \textbf{3.56} & \textbf{4.15} & \textbf{4.62} & \textbf{5.04}  \\ \hline
    \end{tabular}
\end{table}

\begin{figure*}[ht]
    \centering
    \setlength{\abovecaptionskip}{0.cm}
    \includegraphics[ width=7 in]{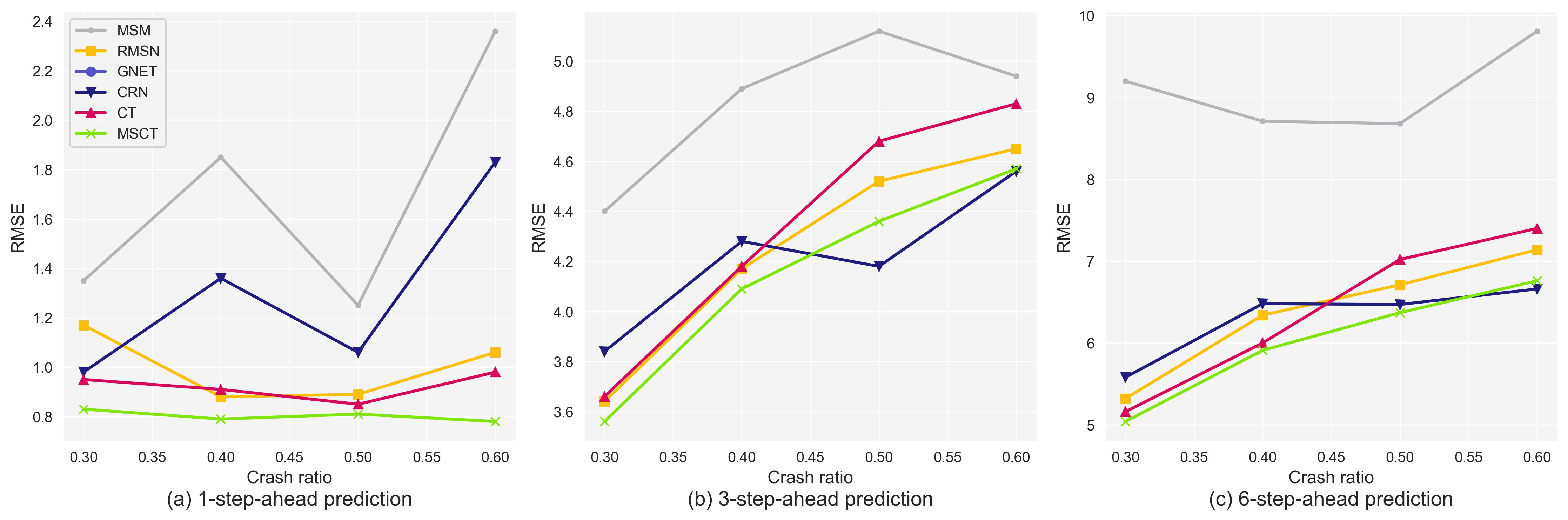}
    \caption{The Impact of Crash Ratio in Real-world Dataset on RMSE}
    \label{fig_8}
\end{figure*}

Considering the influence of the ratio of crash samples in the dataset, we further explore the prediction performance changing with the crash ratio. Fig.\ref{fig_8} shows that for 1-step-ahead prediction, there is no consistent relationship between prediction accuracy and crash ratio. As for other time-step-ahead predictions, there is a slight increase as the crash ratio increases. This reflects the good adaptable capability of these counterfactual models. Because the different distributions of the dataset may cause different selection biases, counterfactual prediction models should be robust for the imbalanced observational dataset. In addition, our proposed model still achieves state-of-the-art performance.

\begin{figure}[!ht]
    \centering
    \setlength{\abovecaptionskip}{0.cm}
    \includegraphics[ width=3.5 in]{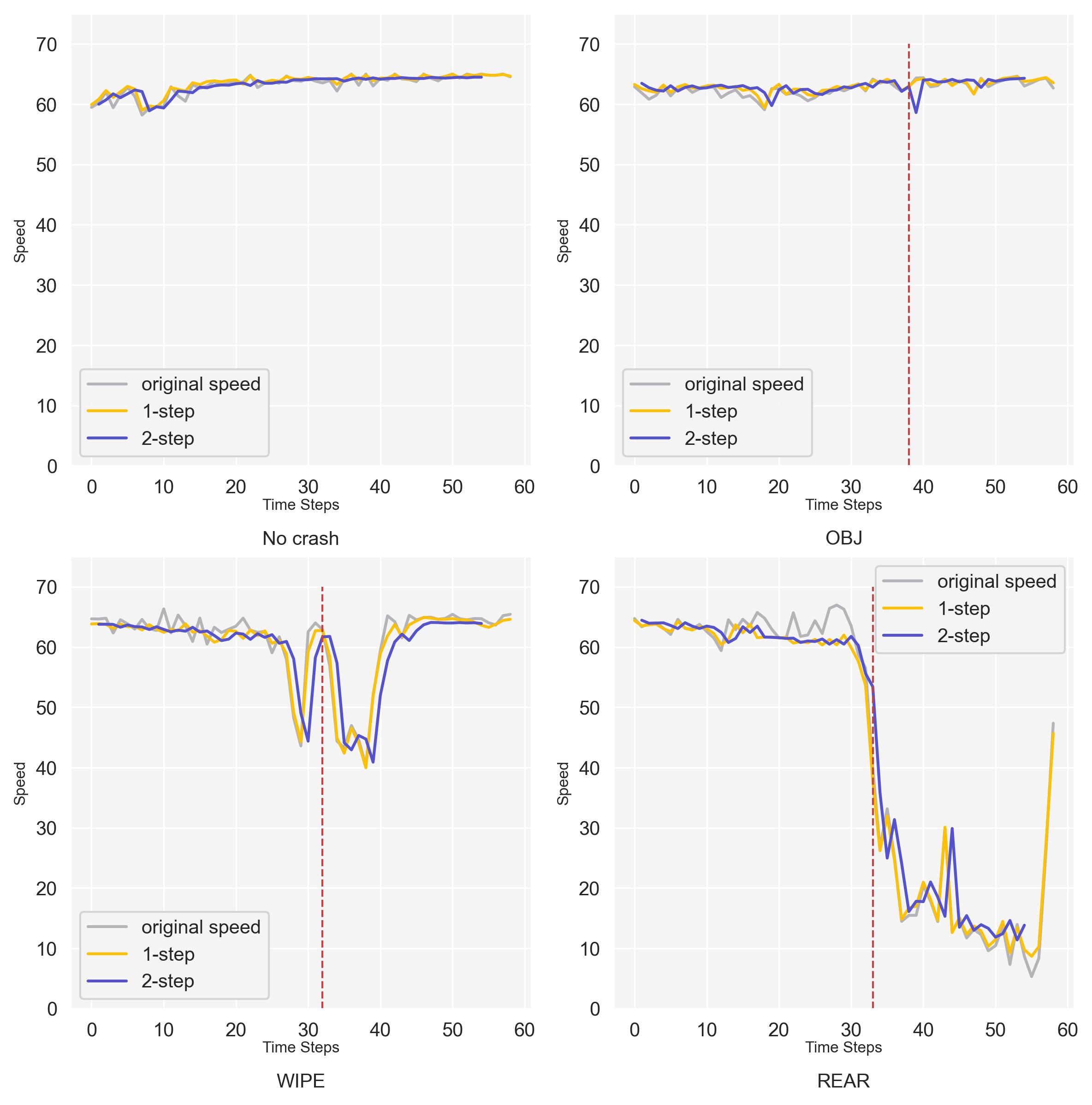}
    \caption{Comparison of original and predicted speed by MSCT across different crash types. The red dotted line indicates the moment of crash}
    \label{fig_9}
\end{figure}

Fig.\ref{fig_9} illustrates the comparison of predicted results by MSCT across different crash types. The four subgraphs demonstrate the various types of crashes: no crash, OBJ, WIPE, and REAR. The 1-step-ahead and 2-step-ahead speeds predicted by the encoder and decoder are compared with the original speed. This figure reveals the varying degrees of impact, and it can be observed that the predicted values closely match the original data for all crash types, demonstrating the ability of MSCT to handle heterogeneous scenarios. 

\subsection{Ablation Study}
To evaluate the effectiveness of each component inside MSCT, the ablation test is conducted using synthetic data. The setting of the ablation study is shown as follows:

\begin{itemize}
    \item \textbf{w/o transf}: replace the transformer module with LSTM network.
    \item \textbf{w/o ps}: remove the PS loss from final loss. Consequently, LSTM pathway in MSCT block and PS head are also removed.
    \item \textbf{w/o bl}: remove balanced loss (i.e., PS loss and HPS loss) from final loss. Consequently, LSTM pathway, PS head, and HPS head are also removed.
    \item \textbf{w/ gr}: replace domain confusion training by gradient reversal training \cite{li_deep_2018, ganin_domain-adversarial_2016}.
\end{itemize}

TABLE \ref{tab_4} summarizes the ablation study results. Overall, MSCT outperforms all its variants in predicting the speed of almost all time steps. When replacing the transformer with LSTM network, although the performance of 1-step-ahead prediction is better than others, the other results predicted by the decoder are much worse. This indicates that the transformer improves the performance for sequence-to-sequence structure and is more suitable for autoregressive prediction. The variants without PS loss and balance loss present poorer ability than MSCT, which proves the effectiveness of propensity score and historical propensity score for eliminating time-varying confounding bias. When using gradient reversal instead of domain confusion, there is almost no difference for short-term prediction, however, the performance becomes worse when predicting long-term speed. In all, the results highlight the importance of transformer, propensity score adjustment, and domain confusion training in the MSCT model.

\begin{table}[!h]
    \renewcommand{\arraystretch}{1.2}
    \centering
    \caption{ABLATION RESULTS FOR SYNTHETIC TRAFFIC DATA (\(\omega=5\))}
    \label{tab_4}
    \begin{tabular}{c|c|c c c c c}
    \hline
        ~ & \(\tau=1\) & \(\tau=2\) & \(\tau=3\) & \(\tau=4\) & \(\tau=5\) & \(\tau=6\) \\ \hline
       w/o transf & \textbf{7.88}  & 11.96 & 12.96 & 13.34 & 13.36 & 13.4  \\ 
        w/o ps & 7.98 & 10.89 & 11.97 & 12.24 & 12.66 & 12.55  \\ 
        w/o bl & 8.04 & 10.97 & 12.12 & 12.46 & 12.51 & 12.38  \\ 
        w/ gr & 7.98 & 10.68 & 11.97 & 12.15 & 12.02 & 12.05  \\ \hline
        MSCT & 7.96 & \textbf{10.47 }& \textbf{11.77} & \textbf{11.99} & \textbf{11.86} & \textbf{11.86}  \\ \hline
    \end{tabular}
\end{table}

\section{Conclusions}
This study has presented a novel causal deep learning model, named Marginal Structural Causal Transformer (MSCT), for counterfactual traffic speed prediction under the intervention of traffic crashes. The proposed model effectively addresses the challenge of time-varying confounding bias in time-series traffic data and the heterogeneous effects of crashes. Due to the lack of ground-truth data, we also propose a synthetic traffic data generation procedure that mimics the causal relationship between crashes and traffic speed. Our experimental results demonstrate that the MSCT outperforms existing models both in synthetic data and real-world data, particularly in longer prediction horizons. The findings highlight the importance of incorporating causal theory into deep learning models for counterfactual outcome prediction in transportation research. In the future, we plan to explore spatial causal relationships as well, considering that traffic data from the road network exhibits spatial dependencies. Besides, the model architecture will be further refined to explore the model's applicability in different transportation scenarios. Overall, MSCT represents the first attempt at counterfactual prediction by deep learning under traffic crash conditions and shows great potential for improving traffic management and decision-making processes in real-world scenarios.

\bibliographystyle{IEEEtran}
\bibliography{IEEEabrv,reference}

\end{document}